\documentclass[10pt,twocolumn,letterpaper]{article}
\usepackage{cvpr,times,epsfig,graphicx,subfigure,braket,authblk,amsmath,amssymb,booktabs,multirow}
\usepackage{array,color,bm,ifthen,tabu,xcolor,colortbl,dblfloatfix}
\usepackage[sort]{cite}
\usepackage[linesnumbered,ruled]{algorithm2e}
\usepackage[pagebackref=true,breaklinks=true,letterpaper=true,colorlinks=true,bookmarks=false]{hyperref}
\cvprfinalcopy

\def\cvprPaperID{***}

\definecolor{darkred}{RGB}{180,50,20}
\definecolor{darkgreen}{RGB}{20,180,50}
\definecolor{purple}{RGB}{90,20,90}

\begin{document}

\title{Learning Instance Segmentation by Interaction}
\author{Deepak Pathak\thanks{Denotes equal contribution} , \ Yide Shentu$^\ast$, \ Dian Chen$^\ast$, \ Pulkit Agrawal$^\ast$, \\
\vspace{-2mm}
Trevor Darrell,  \ Sergey Levine, \ Jitendra Malik \\
\vspace{2mm}
University of California, Berkeley\\
\vspace{2mm}
\small\texttt{\{pathak,fredshentu,dianchen,pulkitag,trevor,slevine,malik\}@berkeley.edu}}

\maketitle
\begin{abstract}
We present an approach for building an active agent that learns to segment its visual observations into individual objects by interacting with its environment in a completely self-supervised manner. The agent uses its current segmentation model to infer pixels that constitute objects and refines the segmentation model by interacting with these pixels. The model learned from over 50K interactions generalizes to novel objects and backgrounds.
To deal with noisy training signal for segmenting objects obtained by self-supervised interactions, we propose robust set loss. A dataset of robot's interactions along-with a few human labeled examples is provided as a benchmark for future research. We test the utility of the learned segmentation model by providing results on a downstream vision-based control task of rearranging multiple objects into target configurations from visual inputs alone. Videos, code, and robotic interaction dataset are available at~\url{https://pathak22.github.io/seg-by-interaction/}.
\end{abstract}

\section{Introduction}
Objects are a fundamental component of visual perception. How are humans able to effortlessly reorganize their visual observations into a discrete set of objects is a question that has puzzled researchers for centuries. The Gestalt school of thought put forth the proposition that humans use similarity in color, texture and motion to group pixels into individual objects~\cite{wertheimer1938laws}. Various methods for object segmentation based on color and texture cues have been proposed~\cite{felzenszwalb2004efficient,carreira2012cpmc,arbelaez2014multiscale,krahenbuhl2014geodesic,isola2014crisp}. These approaches are, however, known to over-segment multi-colored and textured objects.

The current state of the art overcomes these issues by making use of detailed class-specific segmentation annotations for a large number of objects in a massive dataset of web images
~\cite{he2017mask,pinheiro2015learning,long2014fully,hariharan2014simultaneous}. A typical system first uses 1M human annotated Imagenet~\cite{imagenet} images to pretrain a deep neural network. This network is then finetuned using over 700K object instances belonging to eighty semantic classes from the COCO dataset~\cite{lin2014microsoft}. Such data is laborious and extremely time consuming to collect. Furthermore, current systems treat segmentation as an end goal, and do not provide a mechanism for correcting mistakes in downstream tasks. In contrast, one of the main challenges an active agent faces in the real world is adapting to previously unseen scenarios, where recovering from mistakes is critical to success.

\begin{figure}[t!]
    \centering
    \includegraphics[width=\linewidth]{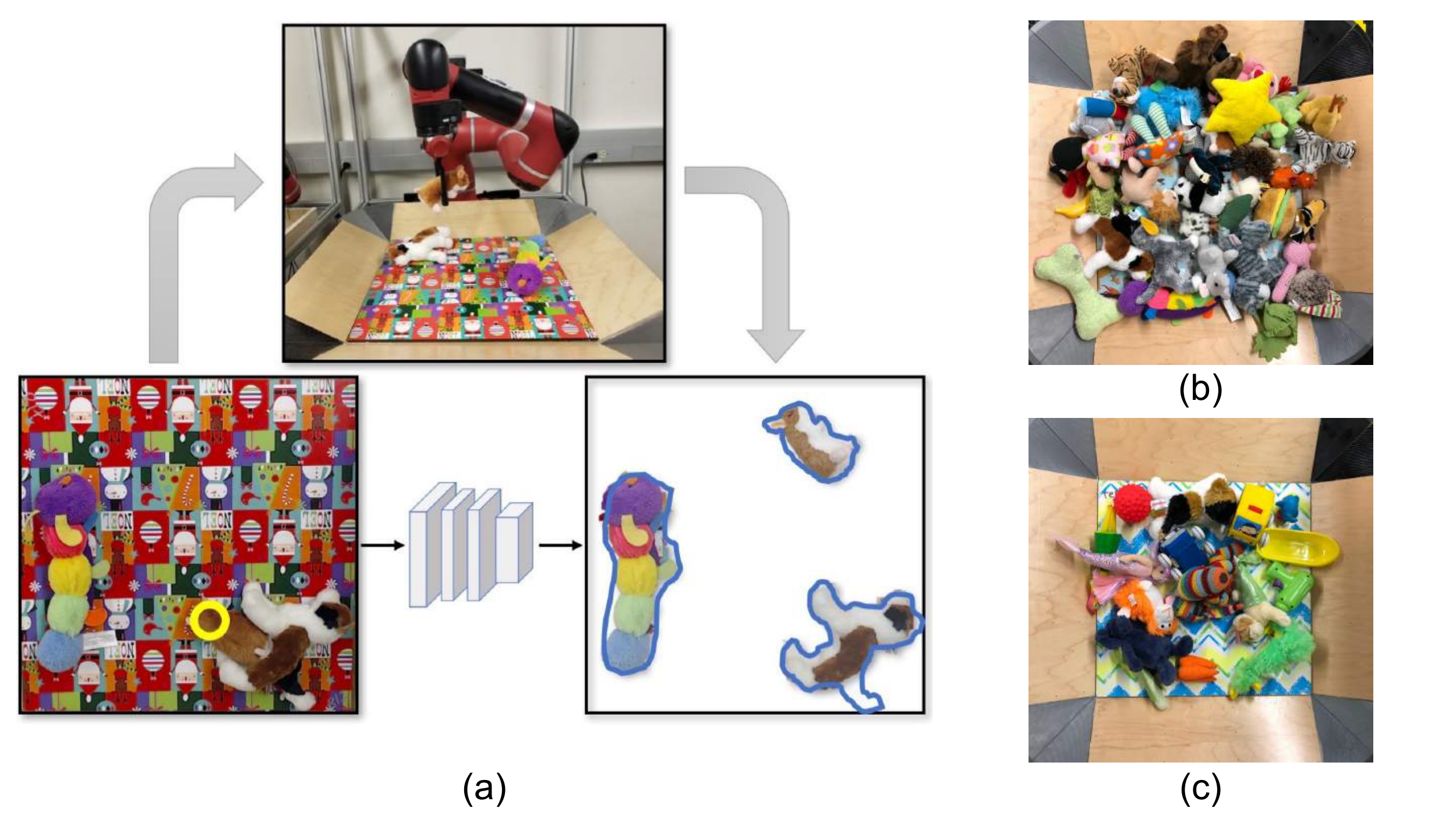}
    \caption{(a): Overview of our approach: a robotic agent conducts experiments in its environment to learn a model for segmenting its visual observation into individual object instances. Our agent maintains a belief about what groups of pixels  might constitute an object and actively tests its belief by attempting to grasp this set of pixels (for e.g. attempts a grasp at the location shown by the yellow circle). Interaction with objects causes motion, whereas interaction with background results in no motion. This motion cue is utilized by the agent to train a deep neural network for segmenting objects. (b),(c): Visualization of the set of thirty six objects used for training (b) and sixteen objects used for testing (c). Validation objects can be seen in supp. materials. Separate sets of backgrounds were used for training, validation and testing.}
    \label{fig:fig1}
    \vspace{-2mm}
\end{figure}

Instead of treating segmentation as a passive process, in this work, our goal is to equip the learner with the ability to actively probe its environment and refine its segmentation model.
One of the main findings in developmental psychology is that, very early on in development, infants have a notion of objects and they expect objects to move as wholes on connected paths, which in turn guides their perception of object boundaries~\cite{spelke1990principles,spelke2007core}. While at first entities merely separated by boundaries might all be the same for an infant, through interaction it is possible for the infant to learn about properties of individual entities and correlate these properties with visual appearance. For example, it is possible to learn that spherical objects roll, a smaller object can be contained inside a larger one, objects with rugged surfaces are harder to push etc. This progression of knowledge starting from delineating the visual space into discrete entities to learning about their detailed physical and material properties naturally paves the path for using this representation for control and eventually categorizing different segmented wholes into different ``object classes.''

In this work, we take the first step towards putting this developmental hypothesis to test and investigate if it is possible for an active agent to learn class agnostic instance segmentation of objects by starting off with two assumptions: (a) there are objects in the world; (b) principle of common fate~\cite{wertheimer1938laws}, i.e. pixels that move together, group together. To that end, we set up an agent, shown in Figure~\ref{fig:fig1}, to interact with its environment and record the resulting RGB images. The agent maintains a belief about how images can be decomposed into objects, and actively tests its belief by attempting to grasp potential objects in the world. Through such self-supervised interaction, we show that it is possible to learn to segment novel objects kept on textured backgrounds into individual instances. We publicly release the collected data (i.e. over 50K interactions recorded from four different views) along with a set of 1700 human labelled images containing 9.3K object segments to serve as a benchmark for evaluating self-supervised, weakly supervised or unsupervised class agnostic instance segmentation method~\footnote{Details at~\url{https://pathak22.github.io/seg-by-interaction/}}.

While interaction is a natural way for an agent to learn, it turns out that training signal for segmentation obtained via self-supervised interactions is very noisy as compared to object masks marked by human annotators.
For example, in a single interaction, the agent might move two nearby objects, which would lead it to mistakenly think of these two objects as one. Dealing with such noise requires the training procedure to be robust, analogous to how in regression, we need to be robust to outliers in the data. However, direct application of pixel-wise robust loss is sub-optimal because we are interested in a set-level statistic such as the similarity between two sets of pixels (e.g. ground-truth and predicted masks) measured for instance using Jaccard index. Such a measurement depends on all the pixels and therefore requires one to define a robust loss over a set of pixels. In this work, we propose a technique, ``robust set loss", to handle noisy segmentation training signal, with the general idea being that the segmenter is not required to predict exactly the pixels in the candidate object mask, rather that the predicted pixels as a set have a good Jaccard index overlap with the candidate mask. We show that robust set loss significantly improves segmentation performance and also reduces the variance in results.

We also demonstrate that the learned model of instance segmentation is useful for visuo-motor control by showing that our robot can successfully re-arrange objects kept on a table into a desired configuration using visual inputs alone. The utility of the learned segmentation method for control shows that it can guide further learning about properties of these segments in a manner similar to how human infants learn about physical and material object properties.
An overview of our approach is shown in Figure~\ref{fig:fig1}.

\section{Related Work}
Our work draws upon the ideas from the active perception~\cite{gibson1979ecological,bajcsy1988active,aloimonos1988active,bajcsy2016revisiting} to build a self-supervised object segmentation system. Closest to our work is ~\cite{pathak2016learning} that makes use of optical flow to generate pseudo ground truth masks from passively observed videos.
We discuss the similarities and differences from past work below.

\vspace{2mm}
\noindent\textbf{Interactive Segmentation:}
Improving the result of segmentation by interaction has drawn a lot of interest ~\cite{fitzpatrick2003first,kenney2009interactive,hausman2015interactive,van2014probabilistic,pajarinen2015decision,bjorkman2010active,nalpantidis2012yes}. However, most these works are concerned with using interaction to segment a specific scene. In contrast, our system uses interactions to actively gather supervision to train a segmentation system that can be used to segment objects in new images. The recent work on SE3 nets~\cite{byravan2017se3} learns to segment and model dynamics of rigid bodies in table-top environments containing boxes. As opposed to using depth data, we show object segmentation results from purely RGB images in visually more complex environment.

\begin{figure*}[t]
\vspace{-2mm}
\centering
\includegraphics[width=.9\linewidth]{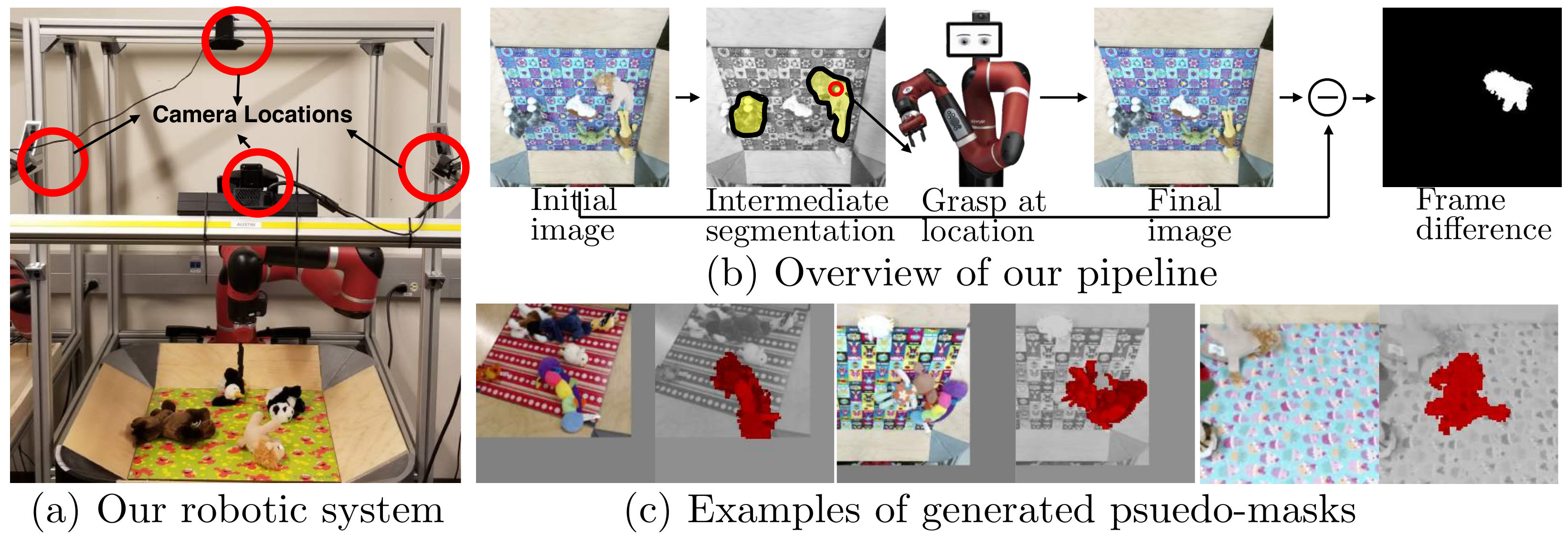}
\caption{Overview of experimental setup and method: (a) Sawyer robot's interactions with objects placed on an arena are recorded by four cameras. The arena is designed to allow easy modification of background texture. (b) From its visual observation (initial image) the robot hypothesizes what group of pixels constitute an object (intermediate segmentation hypothesis). It randomly chooses to interact with one such group by attempting to grasp and place it to a different location on the arena. If the grasped group indeed corresponds to an object, the mask of the object can be obtained by computing the difference image between the image after and before the interaction. The mask obtained from the difference image is used as pseudo ground truth for training a neural network to predict object segmentation masks. (c) Sometimes masks produced by this process are good (first image), but they are often imperfect due to movement of multiple objects in the process of picking one object (second image) or creation of false masks due to lighting changes/shadows.}
\vspace{-2mm}
\label{fig:method}
\end{figure*}

\vspace{2mm}
\noindent\textbf{Self-Supervised Representation Learning:}
In this work, we use a self-supervised method to learn a model for segmentation. A number of recent works have studied self-supervision, using signals such as ego-motion~\cite{agrawal2015learning,jayaraman2015learning}, audio~\cite{owens2016ambient}, colorization~\cite{zhang2016colorful}, inpainting~\cite{pathakCVPR16context}, context~\cite{doersch2014context,NorooziECCV2016}, temporal continuity~\cite{wang2015unsupervised}, temporal ordering ~\cite{misra2016unsupervised}, and adversarial reconstruction~\cite{donahue2016adversarial} for learning visual features as an alternative to using human-provided labels. As far as we are aware, ours is the first work that aims to learn to segment objects using self-supervision from active robotic interaction.

\vspace{2mm}
\noindent\textbf{Self-Supervised Robot Learning:}
Many recent papers have investigated use of self-supervised learning for performing sensorimotor tasks. This includes self-supervised grasping~\cite{pinto2015supersizing,levine2016learning, mahler2017dex}, pushing~\cite{agrawal2016learning,finn2017deep,pinto2016curious,byravan2017se3}, navigation~\cite{pathak2018zero,gandhi2017learning} and rope-manipulation~\cite{nair2017combining,pathak2018zero}.
However, the focus of these works was geared for an end task. Our goal is different -- it is to a learn robust ``segmentation" from noisy interaction signal. Such segmentation can be a building block for multiple robotic applications.

\section{Experimental Setup}
\label{sec:setup}
Our setup, shown in Figure~\ref{fig:method}, consists of a Sawyer robot interacting with objects kept on a flat wooden arena. The arena is observed by
four cameras placed at different locations around it. For diversifying the environment, we constructed the arena in a manner that the texture of the arena's surface could be easily modified. At any point in time, the arena contained 4 to 8 objects randomly sampled from a set of 36 training objects. We set up the agent to interact autonomously with objects without any human supervision. The agent made on average three interactions per minute using the \textit{pick and place} primitive. We used the pick and place primitive as the primary mechanism for interaction as it leads to larger displacement of objects in comparison to say push actions and thereby provides more robust data for learning instance segmentation. We now describe each part of our methodology in detail.

\vspace{2mm}
\noindent\textbf{Pick and Place Primitive:}
The pick action was parameterized by the location (a 2D point on the planar surface of the arena) and rotation of the agent's end effector (i.e., the gripper). The agent approached the pick location from the top with its gripper perpendicular to the arena, rotated by the desired angle and kept wide open. At the pick location, the gripper was closed by a pre-fixed amount to grasp the object (if any). After the grasp, the gripper was moved to the \textit{place} location and opened. If the gripper held an object, the place action caused the object to drop on the arena.
This pick and place motion of the robot was enabled by calibrating the robot's internal coordinate system with the arena kept in front of it using Kinect sensing.

During the process of pick and place, three images (size 350x430 pixels) of the arena were captured: $I_t$ before the pick action, $I'_{t+1}$ when the grasped object is picked but not placed on the arena and $I_{t+1}$ taken after placing the object. All images were captured by positioning the agent's arm in a manner that did not obstruct the view of the arena from any of the four cameras. Note that every pick and place action did not lead to displacement of an object because: (a) either the pick operation was attempted at a location where no object was present (and in this case, $I_t$, $I'_{t+1}, I_{t+1}$ had objects in the same configuration) or (b) the grasping failed and the object was not picked, but was possibly displaced due to contact with the gripper. In case of (b), $I_t$ and $I'_{t+1}$ typically had objects in slightly different configurations whereas $I'_{t+1}$ and $I_{t+1}$ had objects in the same configuration.

\vspace{2mm}
\noindent\textbf{Interaction Procedure:}
Let the agent's current observation be $I_t$ and its belief about group of pixels that constitute an object be $\{s_1^t, s_2^t ..s_K^t \}$, where $s_i^t$ (a binary mask) indicates the set of pixels that belong to the $i^{th}$ group among a total of $K$ groups. The agent interacts to verify if \mbox{$s_j^t (j \in [1, K])$} constitutes an object by attempting to pick and then place $s_j^t$ at a randomly chosen location on the arena. If pixels in $s_j^t$ move, it confirms the agent's belief that $s_j^t$ is an actual object. Otherwise, the agent revises its belief to the contrary. Note that our goal is to show that we can obtain good instance segmentation by interaction, and hence we use standard motion planning procedure~\cite{cowley2013perception,gong2011multi} to simply hard-code the interaction pipeline. Details are in the supplementary.

Since objects were only moved by agent's interactions, the collected data is expected to be highly correlated. To prevent such correlations, the agent executed \textit{a fully automatic reset} after every 25 interactions without any human intervention. In such a reset procedure, the agent moved its gripper from eight different points uniformly sampled on the boundaries of the arena to the center to randomly displace objects. To further safeguard against correlations, the background was periodically changed. Overall, the agent performed more than 50,000 interactions.

\begin{algorithm}[t]
    Pre-train network with passive unsupervised data\\
    \For{\text{iteration t $=$ 1 to T}}
        {
        Record current observation $I_t$\\
        Generate object hypothesis: $\set{s_1^t,\dots s_K^t}$ $\Leftarrow$ CNN$(I_t)$\\
        Randomly choose one hypothesis $s_j^t\in\set{s_1^t,\dots s_K^t}$\\
        Interact with hypothesized object ($move(s_j^t)$)\\
        Record observation $I_{t+1}$\\
        mask $\Leftarrow$ frame$\_$difference($I_t, I_{t+1}$)\\
        \eIf {mask is empty}
            {
                $\{$(x,y), mask, $I^t\}$ is negative training
                example\\
            }
            {
                $\{$(x,y), mask, $I^t\}$ is positive training example\\
            }
        \If {t $\%$ update$\_$interval $==$ 0}
            {
               Update CNN using positive/negative examples
            }
        }
    \caption{Segmentation by Interaction}
    \label{alg:algo1}
\end{algorithm}

\section{Instance Segmentation by Interaction}
\label{sec:method}

The primary goal of this work is to investigate if it is possible for an active learner to separate its visual inputs into individual foreground objects (i.e., obtain instances) and background by self-supervised active interaction instead of human supervision. Broadly the agent moves hypothesized objects and this motion is used to generate (pseudo ground-truth) object masks that are used to supervise learning of the segmentation model.

The major challenge in training a model with such self-generated masks is that they are far from perfect (Figure~\ref{fig:method}). Typical error modes include: (a) false negatives due to complete failure to grasp an object; (b) failure in grasping that slightly perturb the object resulting in incomplete masks; (c) in case two objects are located near each other, picking one object moves the other one, resulting in masks that span multiple objects; (d) erroneous masks due to variation in lighting, shadows and other nuisance factors. Any method attempting to learn object segmentation from interaction must deal with such imperfections in the self-generated \textit{pseudo ground truth} masks.

When near-perfect human annotated masks are available, it is possible to directly optimize per-pixel loss determining whether the pixel belongs to background or foreground. With noisy masks it is desirable to optimize a robust loss function that only forces the predictions to approximately match the noisy ground truth. Since noise in segmentation masks is a global property of the image, it is non-trivial to employ pixel-wise robust loss. We discuss this challenge in more detail and a solution to it by proposing \textit{Robust Set Loss} in section~\ref{sec:robust}.

While there are many methods in the literature for making use of object masks for training instance segmentation systems, without any loss of generality in this work we use the state-of-art method known as DeepMask~\cite{pinheiro2015learning} to train a deep convolution neural network initialized with random weights (i.e., \textit{from scratch}). Note that the use of this method for training CNN is complementary to our contribution of learning from active interaction and dealing with challenges of noisy training signal using \textit{robust set loss}. The DeepMask framework produces class agnostic instance segmentation with the help of two sub-modules: a scoring network and a mask network. The basic idea is to scan image patches at multiple scales using the sliding window approach, and each patch is evaluated by the \textit{scoring network} to determine whether the center pixel of the patch is part of foreground or background. If the center pixel of the image crop belongs to the foreground (i.e., the patch is believed to contain an object), it is passed into the \textit{mask network} to output the segmentation mask. We use the active interaction data generated by the agent to train the \textit{mask} and \textit{scoring} networks.

\subsection{Training Procedure}
The training procedure is summarized in Algorithm~\ref{alg:algo1}. Let the current image observed by the agent be $I_t$. The image is first re-sized into seven different scales given by $2^{i*0.25 - 1.25}, i \in [0, 6]$. For each scale, the output of scoring network is computed for image patches of size 192x192 extracted at a stride of 16. All patches that are predicted by scoring network to contain object segment/masks $\{s_1^t, s_2^t ..s_K^t \}$.

The agent randomly decides to interact with one these object segment hypotheses (say $s_j^t$) using the pick and place primitive described in section~\ref{sec:setup}. For ascertaining if $s_j^t$ indeed corresponds to the object, we compute the difference image $d'_t = I'_{t+1} - I_{t}$. For increasing robustness to noise we only compute the difference in a square region of size 240x240 pixels around the point where robot attempted the pick action. Additional computations to increase robustness of difference image are described in the supplementary materials.

From the difference image, we extract a single mask of connected pixels (say $m_j^t$). If the number of non-zero pixels in this mask are greater than 1000, we regard the pick interaction to have found an object (i.e., the image patch is considered to be a positive example for the \textit{scoring network}). The correspond mask $m_j^t$ is used as training data point for the mask network. Otherwise, we regard the  $s_j^t$ to be a part of the background (i.e., negative example for the \textit{scoring network}). We generate additional training data by repeating the same process for image pairs, $I'_{t+1}$ and $I_{t+1}$ (see section~\ref{sec:setup}).

Furthermore, to account for variance in object sizes, we augment the positive data points by randomly scaling images in the range of [$2^{-0.25}, 2^{0.25}]$ and obtain hard negatives by jittering the positive image patches by more than 64 pixels in L1 distance (i.e. combined jittering along $x$ and $y$ axes) and randomly jitter negative examples for data augmentation. We used a neural network with a ResNet-18 architecture to first extract a feature representation of the image. This feature representation is fed into two branches that predict the score and the mask each. We use a batch size of 32 and stochastic gradient descent with momentum for training.

\subsection{Robust Set Loss}
\label{sec:robust}
The masks computed by the agent's interaction are quite noisy to train the \textit{mask network} using the standard cross entropy loss that forces the prediction to exactly match the noise in each training data point.
Attempting to fit noise is adversarial for the learning process,
as (a) overfitting to noise would hamper the ability to generalize to unseen examples, and (b) inability to fit noise would increase variance in the gradients and thereby make training unstable.

The principled approach of learning with noisy training data is to use a robust loss function for mitigating the effect of outliers.
Robust loss functions have been extensively studied in statistics, in particular, Huber loss~\cite{huber1964robust} applied to regression problems.
However, such ideas have mostly been explored in the context of regression and classification for modeling independent outputs.
Unfortunately, segmentation mask is a ``set of pixels'', where a statistic of interest such as the similarity between two sets of pixels (e.g., ground-truth and predicted masks) measured for instance using Jaccard Index (i.e., intersection over union (IOU)) depends on all the pixels. The dependence of the statistic on a set of pixels makes it non-trivial to generalize ideas such as Huber loss in a straightforward manner. We formulate Robust Set Loss to deal with ``set-level'' noise.

Before discussing the formulation, we describe the intuition behind our formulation using segmentation as an example.
Our main insight is that, if the target segmentation mask is noisy, it is not desirable to force the per-pixel output of the model to exactly match the noisy target. Instead, we would like to impose a soft constraint for only matching a subset of target pixels while ensuring that some  (potentially non-differentiable) metric of interest, such as IOU, between the prediction and the noisy target is greater than or equal to a certain threshold.
In case the threshold is 1, it reduces to exactly fitting the target mask. If the threshold is less than 1, it amounts to allowing a margin between the predicted and the target mask.

The hope is that we can infer the actual (latent) ground-truth masks by only matching the network's prediction with the noisy target up to a margin measured by a metric of interest such as the IOU. Because the network parameters are optimized across multiple training examples, it is possible that the network will learn to ignore the noise (as it is hard to model) and predict the pattern that is common across examples and therefore easier to learn. The pattern ``common" across examples is likely to correspond to the actual ground truth. We operationalize this idea behind the \textit{Robust Set Loss} (RSL) via a constrained optimization formulation over the output of the network and the noisy target mask.

Consider the pixel-wise labeling of an image $I$ as a ``set'' of random variables $X=\{x_0,\ldots,x_n\}$ where $x_i\in\mathcal{L}$, where n is total number of pixels and $\mathcal{L}: \{0,1\}$ is the set of possible labels that can be assigned to a pixel.
Let the latent ground truth label corresponding to the image $I$ be $P(X|I)$  and the noisy mask collected by interaction be $M_I = f(P(X|I))$, where $f$ is an arbitrary non-linear function. Let the predicted mask be $Q(X|\theta,I)$, where $\theta$ are the parameters of the neural network. We want to minimize the distance between the prediction and latent ground truth measured using KL-divergence, $D\left(P(X)\|Q(X|\theta)\right) = - \mathbb{E}_{X\sim P}\left[\log Q(X|\theta)\right]$. Assuming, the latent target mask is discrete, $P(X=\hat{X})=\Pi_{i}\big[x_i=\hat{x}_i\big]$, where $\hat X$ is the prediction.

Given the network output $Q(X|\theta,I)$ and the noisy label set (i.e., mask collected by interaction) $M_I$, the goal is to optimize for the latent target $P(X|I)$ which is within a desired margin from the noisy mask $M_I$.  The network prediction $Q(X|\theta,I)$ will then be trained to match $P(X|I)$.

We assume that the latent target is a mask with values in the label set $\mathcal{L}$.
Hence, we model it as a delta function $P(X)=[X=\hat{X}]=\Pi_{i}\big[x_i=\hat{x}_i\big]$.
The distance of this latent target from the predicted distribution is measured via KL-divergence which in case of delta function reduces to $D\left(P(X)\|Q(X|\theta)\right) = - \mathbb{E}_{X\sim P}\left[\log Q(X|\theta)\right] = -\sum_{i} \log q_i(\hat{x}_i)$.
The final optimization problem is formulated as follows:
\begin{align}
\underset{\theta,\hat{X},\xi}{\text{minimize}} & & & -\sum_{i} \log q_i(\hat{x}_i) + \lambda^T\xi \nonumber & \\
\text{subject to} & & & \text{IoU}(\hat{X},M_I) \geq b - \xi,\quad \xi \geq 0 & \label{eq:KL}
\end{align}
where $\xi$ is the slack variable.
We optimize the above objective by approximate discrete optimization~\cite{papandreou2015weakly} in each iteration of training.
Details of the optimization procedure are in the supplementary material. The approximate discrete optimization is fast and takes approximately $0.35$ seconds for a batch of 32 examples.

Note that our formulation could be thought of as a generalization of the CCNN constrained formulation proposed in Pathak et. al.~\cite{pathakICCV15ccnn} with several key differences: (a) we handle non-linear, non-differentiable constraints compared to only linear ones in CCNN, (b) we propose a discrete formulation compared to a continous one in CNN, and (c) our main goal is to handle robustness in set data while CCNN's goal is to learn a pixel-wise ground truth from image level tags in weakly supervised segmentation setting. (d) Moreover, our optimization procedure is an approximate discrete solver while CCNN used projected gradient descent, which would be impractical with Jaccard Index like constraints.

\subsection{Bootstrapping the Learning Process Using\\Passive Self-Supervision}
Without any prior knowledge, the agent's initial beliefs about objects will be arbitrary, causing it to spend most of its time interacting with the background. This process would be very inefficient. We address this issue by assuming that initially our agent can passively observe objects moving in its environment. For this purpose we use a prior robotic pushing dataset~\cite{agrawal2016learning} that was constructed by a robot randomly pushing objects in a tabletop environment. We apply the method of~\cite{pathak2016learning} to automatically extract masks from this data, which we use to pre-train our ResNet-18 network (initialized with random-weights). Note that this method of pre-training is completely self-supervised and does not rely on any human annotation, and it is quite natural to combine passive observation and active interaction for self-supervised learning.

\section{Baselines and Comparisons}
\label{sec:baselines}
We compare the performance of our method against a state-of-the-art bottom up segmentation method called Geodesic Object Proposals (GOP)~\cite{krahenbuhl2014geodesic}, and a top-down instance segmentation method called DeepMask~\cite{pinheiro2015learning} which is pre-trained on 1M ImageNet and then finetuned in a class agnostic manner using over 700K strongly supervised masks obtained from the COCO dataset.
We incorporated NMS (non-max suppression) into DeepMask, which significantly boosted its performance on our dataset. In order to reduce the bias from the domain shift of transferring from web images to images recorded by the robot, we further removed very large masks that could not possibly correspond to individual objects from the outputs of both methods.
Even after these modifications, we found GOP to output proposals that corresponded to other smaller parts of the arena such as the corners. We explicitly removed these proposals and dubbed this method as GOP-Tuned.

\vspace{2mm}
\noindent\textbf{Training/Validation/Test Sets:}
We used 24 backgrounds for training, 6 for validation and 10 for testing. We used 36 different objects for training, 8 for validation and 15 for testing. The validation set consisted of 30 images (5 images per background), and the test set included 200 images (20 images per background). We manually annotated object masks in these images for the purpose of evaluation. In addition, we also provide instance segmentation masks for 1470 training images containing 7946 objects to promote research directions looking at combining small quantities of high-quality annotations along with larger amounts of potentially noisy data collected via self-supervision.

\vspace{2mm}
\noindent\textbf{Metric:} The performance of different segmentation systems is quantified using the standard mean average-precision (mAP;~\cite{everingham2015pascal}) metric at different IoU (intersection over union) thresholds. Intuitively, the mAP at an IoU threshold of $th$ counts generated proposals with an IoU $<th$ with the ground truth as a false positive and penalizes the system for each ground truth proposal that is not matched by a predicted proposal with IoU $>th$.
Higher mAP indicates better performance.

\begin{table}[t!]
\centering
\resizebox{\linewidth}{!}{%
    \begin{tabular}{llcc}
        \toprule
        Method & Supervision &AP at IU 0.3& AP at IU 0.5 \\
        \midrule
        GOP & Bottom up & 10.9 & 04.1 \\
        GOP (tuned) & Bottom up & 23.6 & 16.3 \\
        DeepMask & Strong Sup. & 44.5 & 34.3\\
        DeepMask (tuned) & Strong Sup. & 61.8 & 47.3\\
        \midrule
        Ours + Human & Semi-sup. & 43.1 $\pm$ 2.6 & 21.1 $\pm$ 2.6 \\
        \midrule
        Ours & Self-sup. & 41.1 $\pm$ 2.4 & 16.0 $\pm$ 2.6\\
        \textbf{Ours + Robust Set Loss} & Self-sup. & 45.9 $\pm$ 2.1 & 22.5 $\pm$ 1.3\\
        \bottomrule
    \end{tabular}}
\vspace{-0.2mm}
\caption{Quantitative comparison of our method with bottom-up (GOP~\cite{krahenbuhl2014geodesic}), learned top-down (DeepMask~\cite{pinheiro2015learning}) segmentation methods and optimization without robust set loss on the full test set. We report the mean and standard deviation for our approach. Note that our approach significantly outperforms GOP, but is outperformed by DeepMask that uses strong manual supervision of 700K+ COCO segments and 1M ImageNet images. Adding 1470 images (contains 7946 object instances) with clean segmentation masks labeled by humans improves performance of our base system. The robust set loss not only improves the mean performance over normal cross-entropy loss but also decreases the variance by handling noise across examples.}
\label{tab:ap}
\end{table}

\section{Results and Evaluations}
\begin{figure*}[t!]
\centering
\subfigure[\smaller Performance vs. Interactions]{\label{fig:ap3}\includegraphics[width=0.32\textwidth]{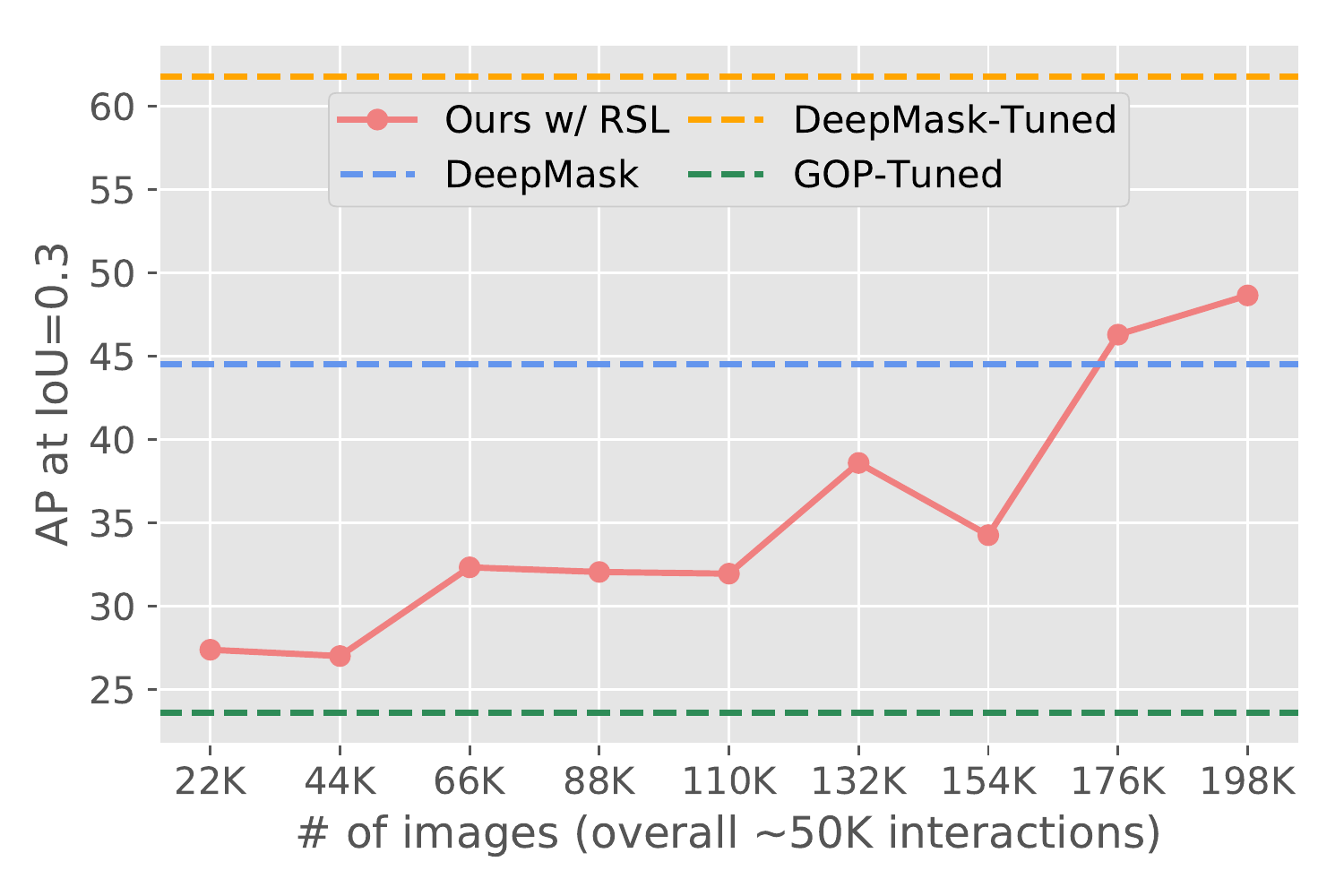}}
\subfigure[\smaller Successes vs. Interactions]{\label{fig:betterdata}\includegraphics[width=0.32\textwidth]{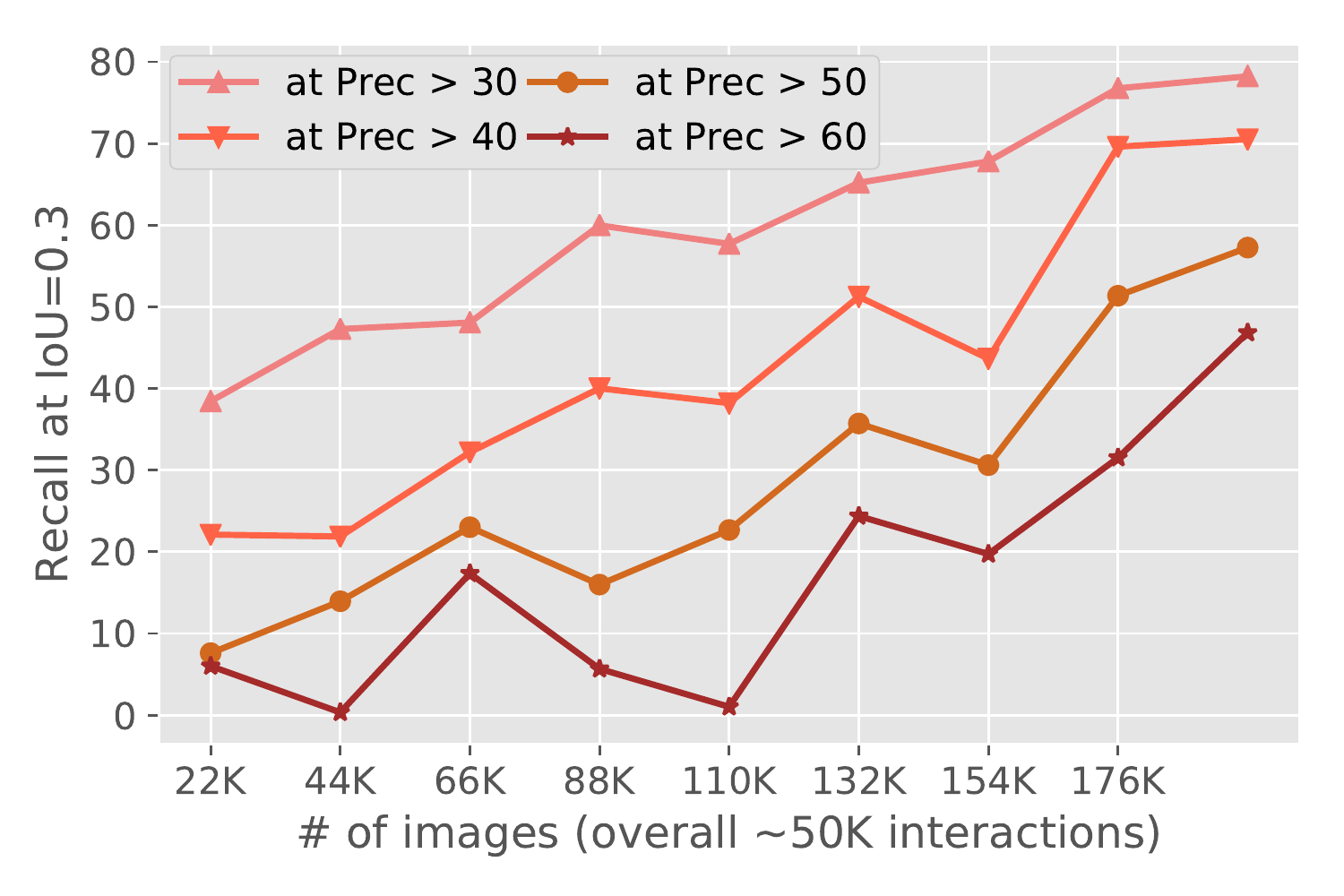}}
\subfigure[\smaller Precision vs. Recall]{\label{fig:precrec}\includegraphics[width=0.32\textwidth]{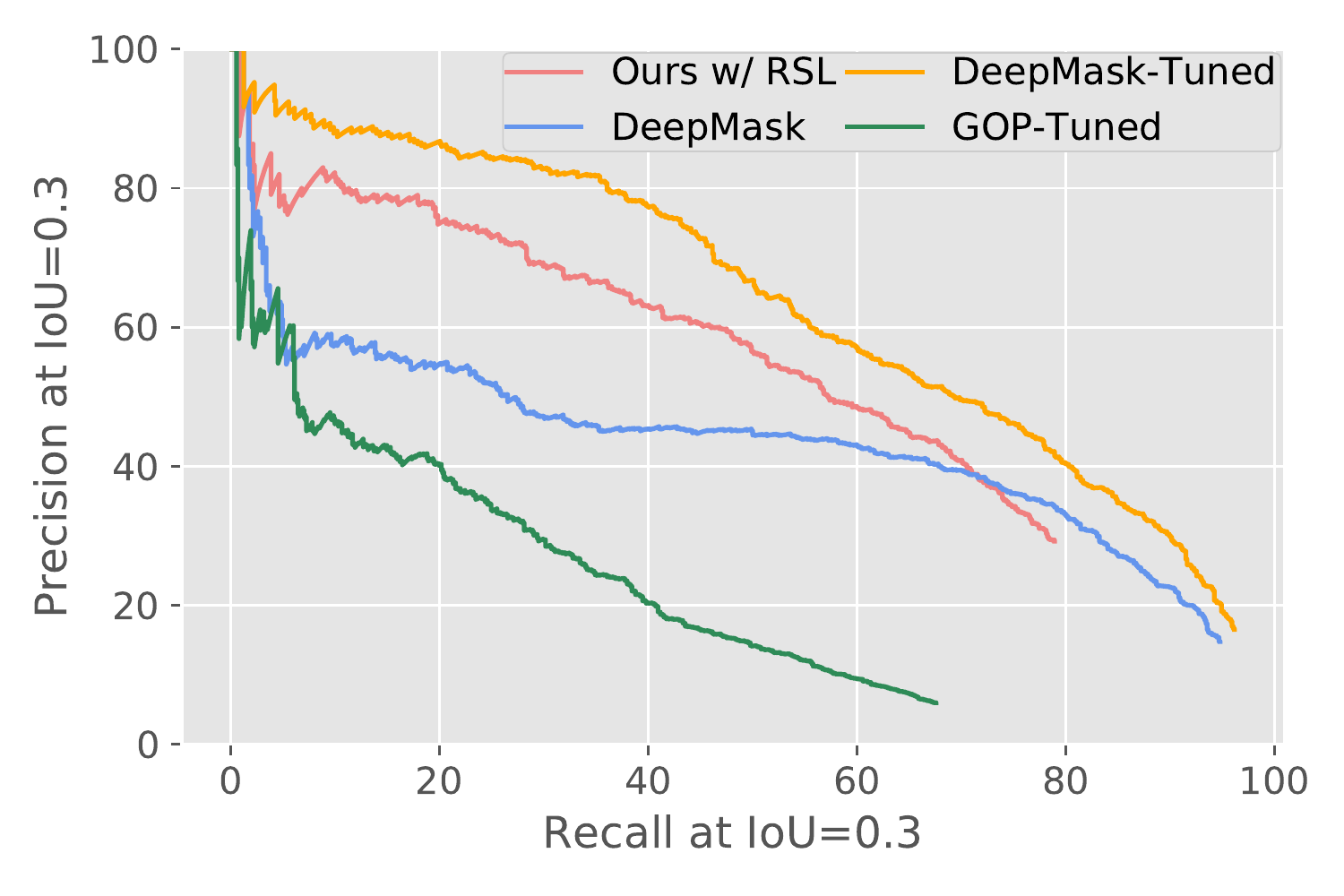}}
\caption{Quantitative evaluation of the segmentation model on the held-out test. (a) The performance of our system measured as mAP at IoU of 0.3 steadily increases with the amount of data. After 50K iterations our system significantly beats GOP tuned with domain knowledge (i.e. GOP-Tuned; section \ref{sec:baselines}). (b) The efficacy of experimentation performed by the robot is computed as the recall of ground truth objects that have IoU of more than 0.3 with the group of pixels that the robot believes to be objects. The steady increase in recall at different precision threshold shows that the robot learns to perform more efficient experiments with time. (c) Precision-Recall curves re-confirm the results.}
\label{fig:performance}
\end{figure*}

We compare the performance of our system against GOP and DeepMask using the AP at IoU 0.3 metric on the held-out testing set as shown in Figure~\ref{fig:ap3}. Our system significantly outperforms GOP even when it is tuned with domain knowledge (GOP-tuned), is superior to DeepMask trained with strong human supervision, but is outperformed when non-max supression (NMS) thresholds and other domain specific tunings are applied  to DeepMask outputs (i.e. DeepMask-tuned). These results are re-confirmed by the precision-recall curves shown in Figure~\ref{fig:precrec}. These results indicate that our approach is able to easily outperform methods relying on hand-engineered bottom-up segmentation cues, but there is still a substantial way to go before matching the performance of a system trained using strong human-supervision (i.e. DeepMask). However, it is encouraging to see from Figure~\ref{fig:ap3} that the performance of our system is steadily increasing with the amount of noisy self-supervised data collected via interactions.

Results in table~\ref{tab:ap} further reveal that adding a few images (i.e. 1470 images containing 7946 object instances) with clean segmentation masks during training (\textit{Ours + Human}) helps improve the performance over our base system possibly due to reduction in noise in training signal. Finally, the robust set loss significantly improves performance.

\begin{figure}[t]
\centering
\includegraphics[width=\linewidth]{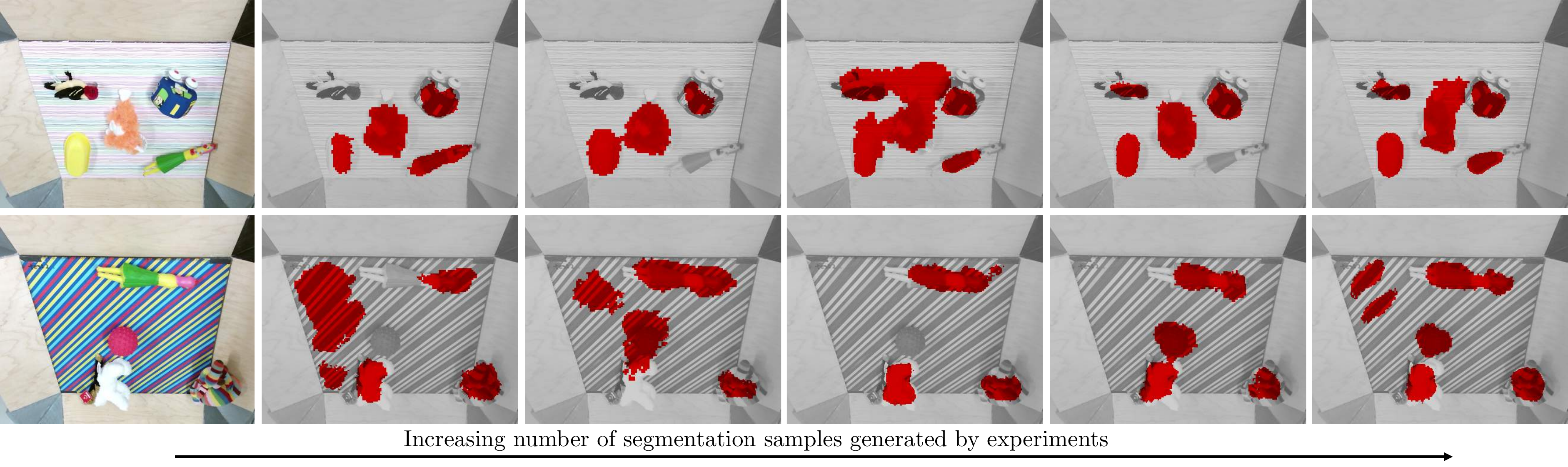}
\caption{The progression of segmentation masks produced by the method as number of experiments conducted by the report increase on held-out test dataset (from left to right). The number of false positives reduce and the quality of masks improve.}
\label{fig:progression}
\end{figure}

While the curves in Figure~\ref{fig:ap3} show an overall increase in performance with increasing amounts of interaction data, there are few intermediate downward deflections. This is not surprising in an active learning system because the data distribution encountered the agent is continuously changing. In our specific case, when backgrounds that are significantly different from the existing training backgrounds are introduced, the existing model has potentially over-fit to previous training backgrounds and the overall performance of the system dips. As the agent interacts in its new environment, it adapts its model and the performance eventually recovers and improves beyond what it was prior to the introduction of the new background. Note that passive learning systems also encounter easy and hard examples, but because the training is batched, in contrast to an active system these examples are uniformly sampled throughout the course of training and therefore such upward/downward fluctuations in performance with increasing amount of data are almost never seen in practice.

\paragraph{Qualitative Comparison}
\begin{figure*}[t]
\vspace{-3mm}
\centering
\includegraphics[width=0.9\linewidth]{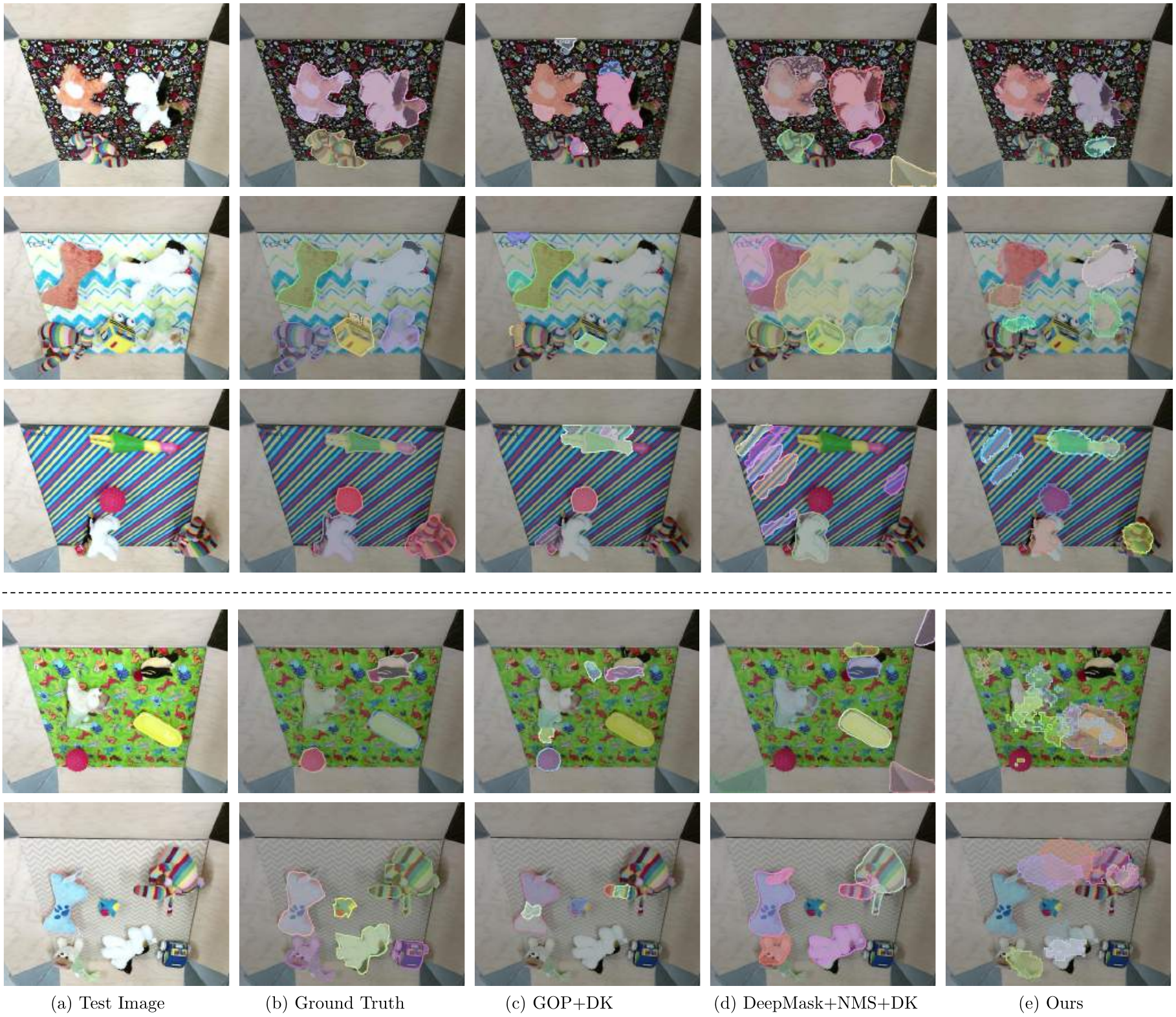}
\caption{Visualization of the object segmentation masks predicted by a bottom up segmentation method GOP with Domain Knowledge (c; GOP tuned), a top down segmentation method trained with strong supervision using more 700K human annotated ground truth object masks (d; DeepMask) and our method (e). The top three rows show representative examples of cases when our methods predicts good masks and the bottom two rows illustrate the failure mode of our method. In general GOP has low recall. The performance of our method and DeepMask is general. Our dominant failure mode is prediction of small disconnected masks (row 4).}
\vspace{-2mm}
\label{fig:seg_vis}
\end{figure*}

Visualization of instance segmentation output of various methods on the test set in Figure~\ref{fig:seg_vis} shows that our method generalizes and can segment novel objects on novel backgrounds. Qualitatively, our method performs similarly to DeepMask, despite receiving no human supervision. The performance of GOP is significantly worse due to low recall.  While in most cases our method produces a connected segmentation mask, in some cases it produces a mask with disconnected jittered pixels (e.g., row-4 in Figure~\ref{fig:seg_vis}). The improvement in the quality of segmentation with agent's experience is visualized in Figure~\ref{fig:progression}. Spurious segmentation of background reduces over time and the recall of objects increases.

\subsection{Active Interactions v/s Passive Data Collection}
\label{sec:better_exp}
An active agent continuously improves its model and therefore not only collects more data, but also higher quality data with time. This suggests that an active agent might require fewer data points than a passive agent for learning.
In case of object segmentation, higher quality of data collected by an agent would be reflected by generation of object hypothesis that have higher recalls at lower false positive rates. We tested if the quality of data generated by active agent improves over time by computing the recall of the ground truth objects using object hypothesis generated by our agent in \emph{novel} environments at different precision thresholds.
Figure~\ref{fig:betterdata} shows that the recall increases over time indicating that our agent learns to perform better experiments with time on the held-out test backgrounds and objects.

\subsection{Analyzing Generalization}
\label{sec:analyze_gen}
Previous results have shown that the performance of our system increases with amount of data. A natural question to ask is, what kind of data would be more useful for agent to learn a segmentation model that will generalize better. In order to answer this question, we investigated whether our system generalized better to new objects or to new backgrounds. For our investigation we constructed four sets of images: (A) training objects on training backgrounds; (B) training objects on test backgrounds; (C) test objects on training backgrounds; and (D) test objects on test backgrounds. If our system generalizes to objects better than background, then changing from training to test objects (but keeping the training backgrounds) should lead to a smaller drop in performance as compared to changing from training to test backgrounds (but keeping the training objects).

When mAP is computed at IoU threshold of 0.3, we find this indeed to be the case. However, when mAP is computed at a threshold of 0.5 we find the reverse trend to hold true. These results suggest that if the quality of mask is not critical (i.e. IoU of 0.3 is sufficient), using larger number of backgrounds is likely to result in better generalization. Alternatively, in use cases where the mask quality is critical, (i.e. IoU of 0.5) using a larger set of objects is likely to result in better generalization.

\begin{figure}[t!]
\centering
\includegraphics[width=0.9\linewidth]{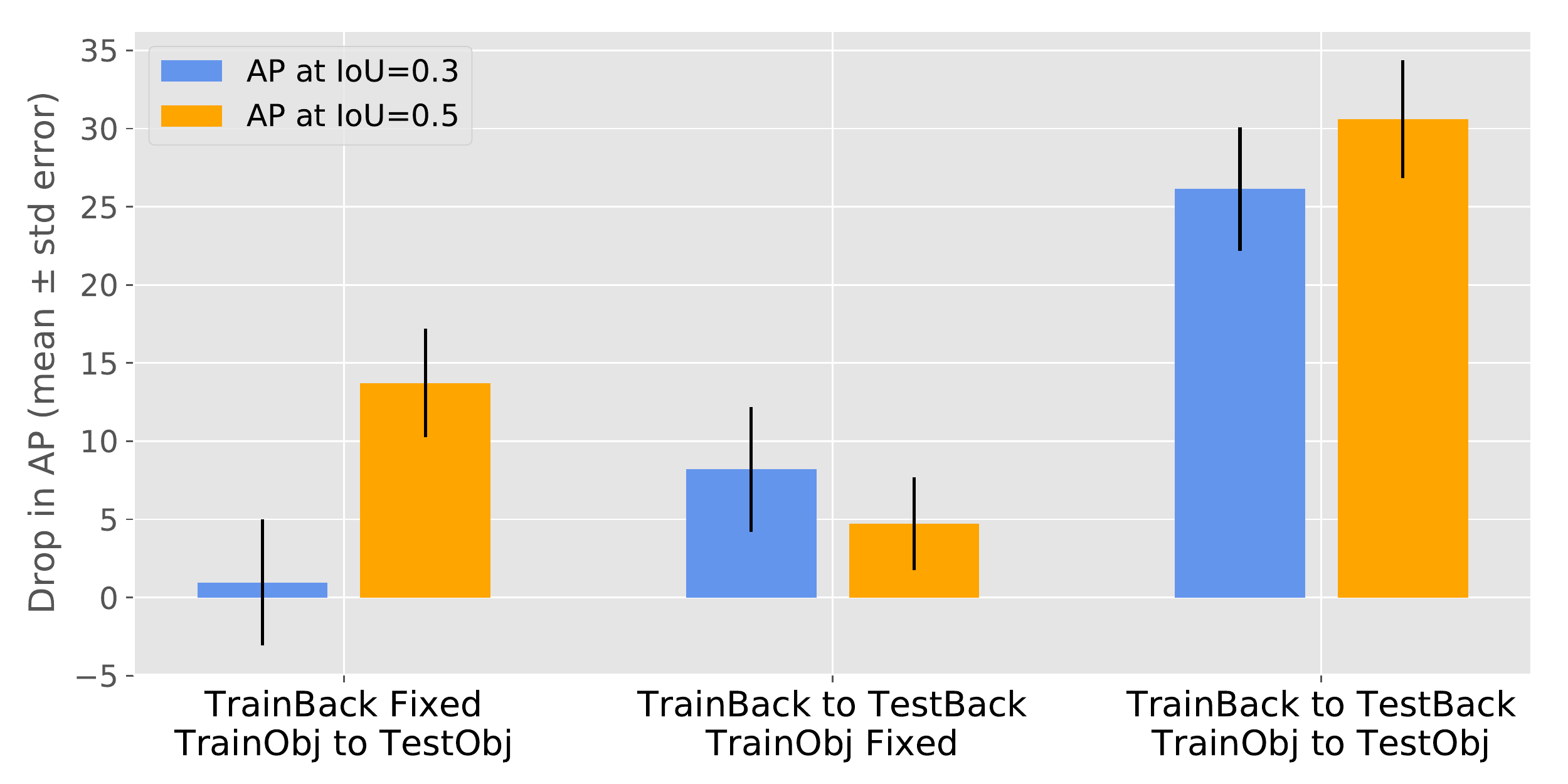}
\caption{\small{Fine-grained generalization analysis of our model. The y-axis denotes the drop in performance (i.e., lower the better) due to the change scenarios specified on the x-axis. When the performance is measured at IoU 0.3, the generalization of our model is better to novel objects as compared to novel backgrounds. However, when the quality of masks is more heavily penalized (i.e. IoU 0.5) the generalization is better to backgrounds as compared to novel objects.}}
\label{fig:generalization}
\vspace{-2mm}
\end{figure}

\subsection{Using Segmentation for Downstream Tasks}
Until now, we have shown results on object segmentation. Next we evaluated if the segmentation returned by our system could be used for downstream tasks by perception or control systems. Towards this end we evaluated performance on the task
rearranging objects kept on a table into a desired configuration.

\subsubsection{Object Rearrangement}
\label{sec:rearrange}

\begin{figure}
\centering
\includegraphics[width=0.9\linewidth]{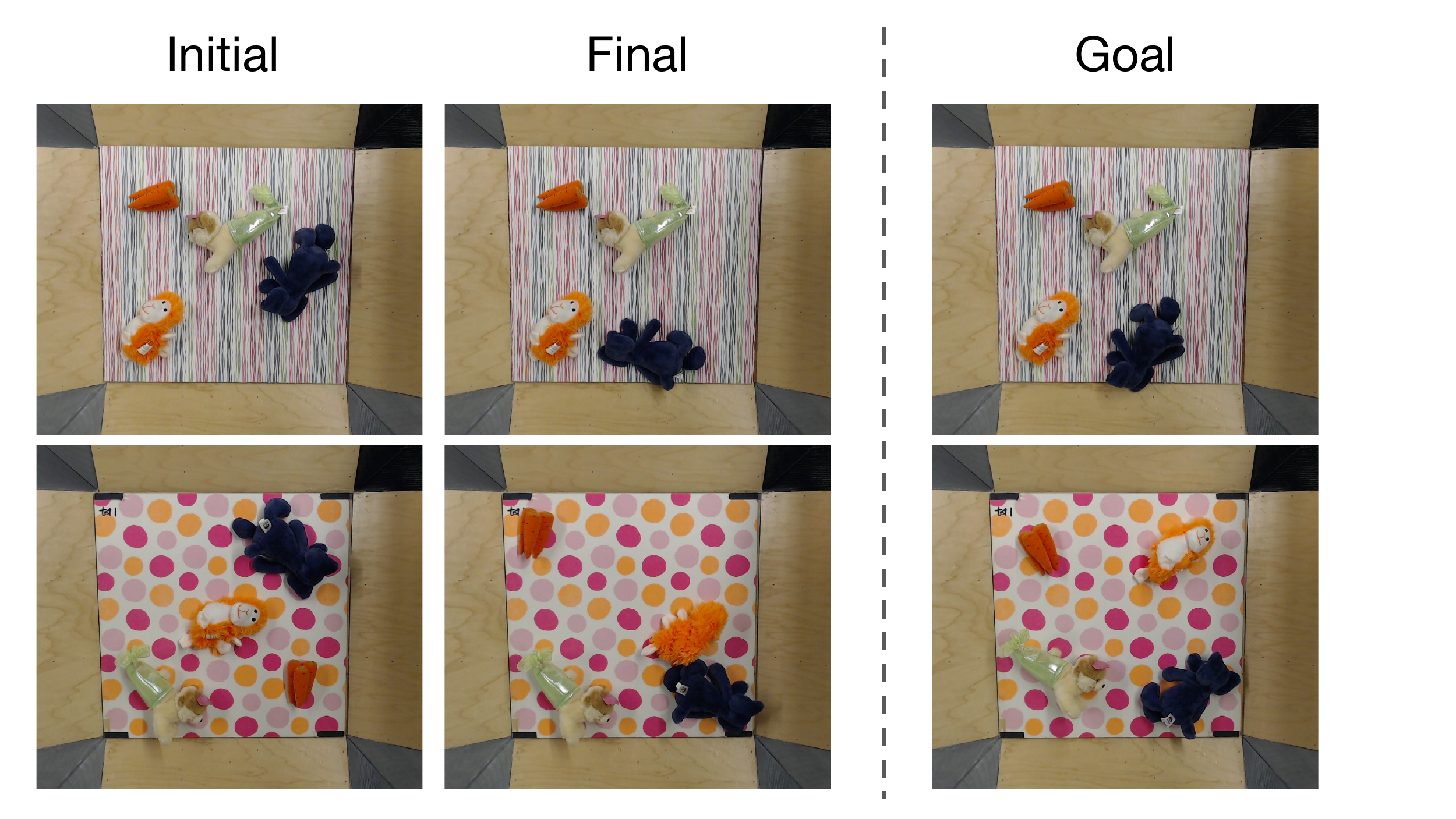}
\includegraphics[width=0.9\linewidth]{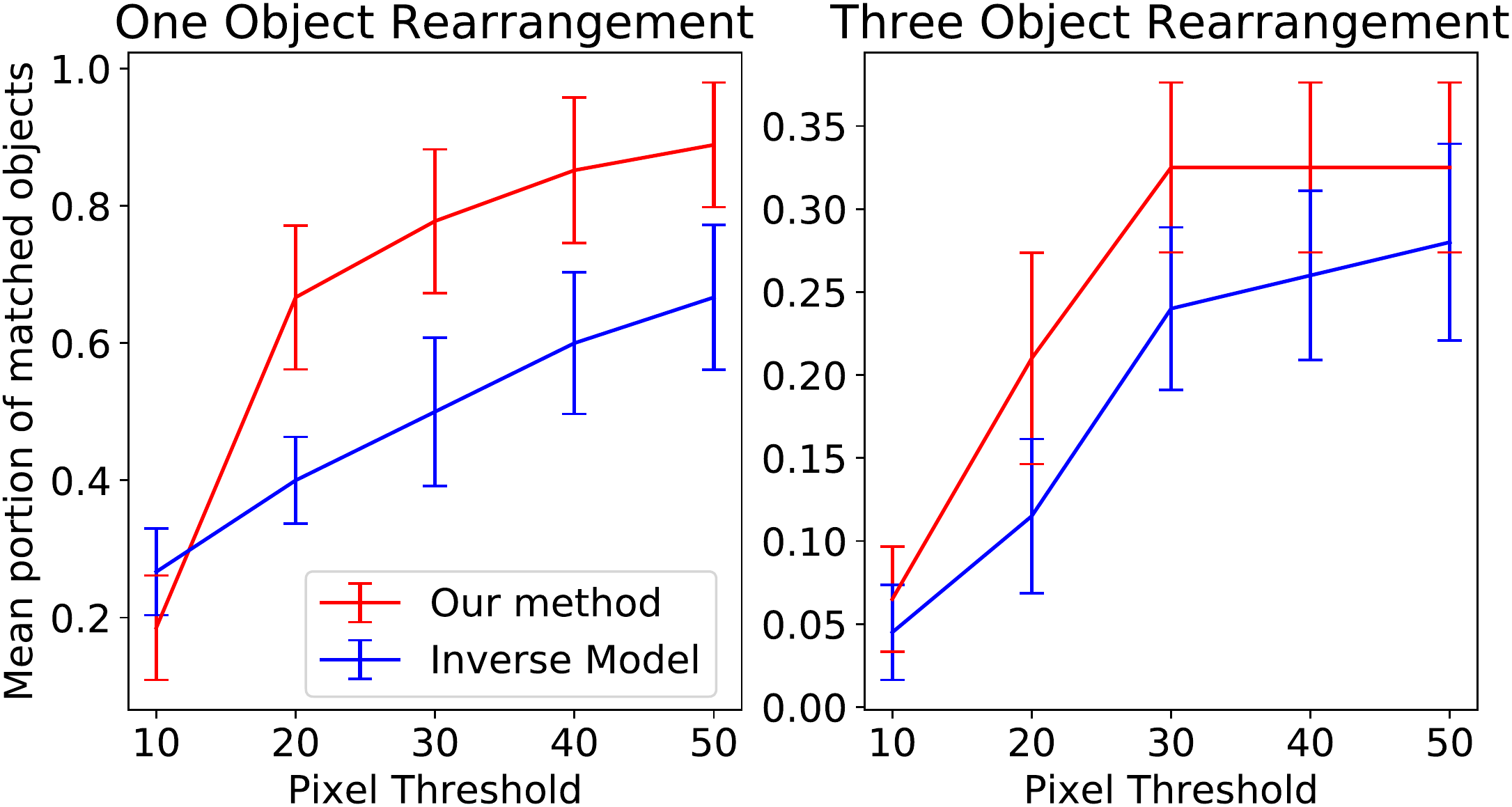}
\caption{\small{The agent was tasked to displace objects in the initial image to their configuration shown in the target image. The final image obtained after manipulation performed by our system is shown in the middle column. Our system made use of the object segmentation learned using active interaction and a hand-designed controller described in section~\ref{sec:rearrange} to perform this task. The majority of failures in our system were due to failures in feature matching or object grasping.}}
\label{fig:rearange}
\end{figure}

We tasked the system to rearrange the objects in its current visual observation into the configuration of objects show in a target image. We did not provide any other information to the agent. 1-3 objects were displaced between the current observation of the agent and the target image. Our robot is equipped with a pick and place primitive as described in section \ref{sec:setup}. If the quality of segmentation is good, it should be possible to match the objects between current and target image and use the pick/place primitive to displace these objects. Since our goal in this work is not to evaluate the matching system, we use off the shelf features extracted from AlexNet trained for classifying Imagenet images.

Our overall pipeline looks as following: (a) obtain a list of object segments produced by our method from current and target image; (b) crop a tight window around the object segments from the original image and pass it into AlexNet to compute feature representation per segment; (c) match the segments between current and target image to determine to what locations the objects in current should be moved to; (d) use the pick/place primitive to move the match objects one by one until the matched objects are within 15 pixels of each other. The robot is allowed a maximum of ten interactions. Qualitative results depicting the performance our system at the rearrangement task are shown in Figure ~\ref{fig:rearange}. While our system is successful sometimes, it fails at many occasions. However most of these failures are a result of failures in feature matching or object grasping. Our system outperforms the inverse model method of ~\cite{agrawal2016learning} for re-arranging objects that doesnot make use of explicit instance segmentation.

\section{Discussion}
In this work, we have presented a method for using active self-supervision to reorganize visual inputs into object instances. The performance of our system is likely to benefit from obtaining better pseudo ground truth masks by the use of better grasping techniques, use of other interaction primitives and joint learning of perceptual and control systems where the interaction mechanism also improves with time.

In order to build general purpose sensorimotor learning systems, it is critical to find ways to transfer knowledge across tasks.
While one approach is to come up with better algorithms for transfer learning, the other is to make use of more structured representations of sensory data than obtained using vanilla feed-forward neural networks. This work, builds upon the second view, in proposing a method for segmenting an image into objects in the hope that object-centric representations might be an important aspect of future visuo-motor control systems.
Our system is only the first step towards the grander goal of creating agents that can self-improve and continuously learn about their environment.

\subsubsection*{Acknowledgments}
We would like to thank members of BAIR community for fruitful discussions. This work was supported in part by ONR MURI N00014-14-1-0671; DARPA; NSF Award IIS-1212798;
Berkeley DeepDrive, Valrhona Reinforcement Learning Fellowship and an equipment grant from NVIDIA. DP is supported by the Facebook graduate fellowship.

{\small
\bibliographystyle{splncs}
\bibliography{main}
}
\clearpage

\appendix
\section{Supplementary Material}
\label{supp}
We evaluated our proposed approach across number of environments and tasks. In this section, we provide additional details about the experimental task setup and hyperparameters.

\subsection{Robust Set Loss}
We optimize the above objective by approximate discrete optimization in each iteration of training.
The approximation is that all the pixels inside and outside the noisy mask $M_I$ are shifted by a common step size.
We reduce the overall problem into an alternating optimization in $P$ and $\theta$. First, the latent $P$ is obtained by taking per-pixel $argmax$ of a distribution $P'$ obtained by modifying per-pixel logits of output $Q$. We shift the logits of the pixels of $Q$, that are inside and outside the noisy mask $M_I$, by a separate bias until $P$ satisfies the IoU constraint. Secondly, the network $Q$ is trained with $P$ as ground truth. This alternating process goes on until the network $Q$ reaches convergence.
The optimization of $P$ in the inner loop is quite fast and takes less than 1ms per image.

\subsection{Interaction Procedure}
The hypothesized object mask ($s_j^t$) is used to determine the pick action of the \textit{pick and place primitive}. A useful heuristic for picking objects is to grasp them perpendicular to their principal axis. We use the object mask to compute the major axis using principle component analysis (PCA). The desired gripper orientation is set to the angle perpendicular to the major axis of the masked pixels and the pick location as the centroid of the mask. The place location is chosen randomly and uniformly across the entire arena.

\subsection{Results for AP at IoU 0.5 }
As discussed in main paper (Figure-4 and Table-1 in the main paper), our method performs very well at AP of IoU 0.3, performing at par with fully supervised DeepMask~\cite{pinheiro2015learning}.

\begin{figure}[ht!]
\centering
\subfigure[\smaller Performance vs. Interactions]{\label{fig:ap5}\includegraphics[width=0.32\textwidth]{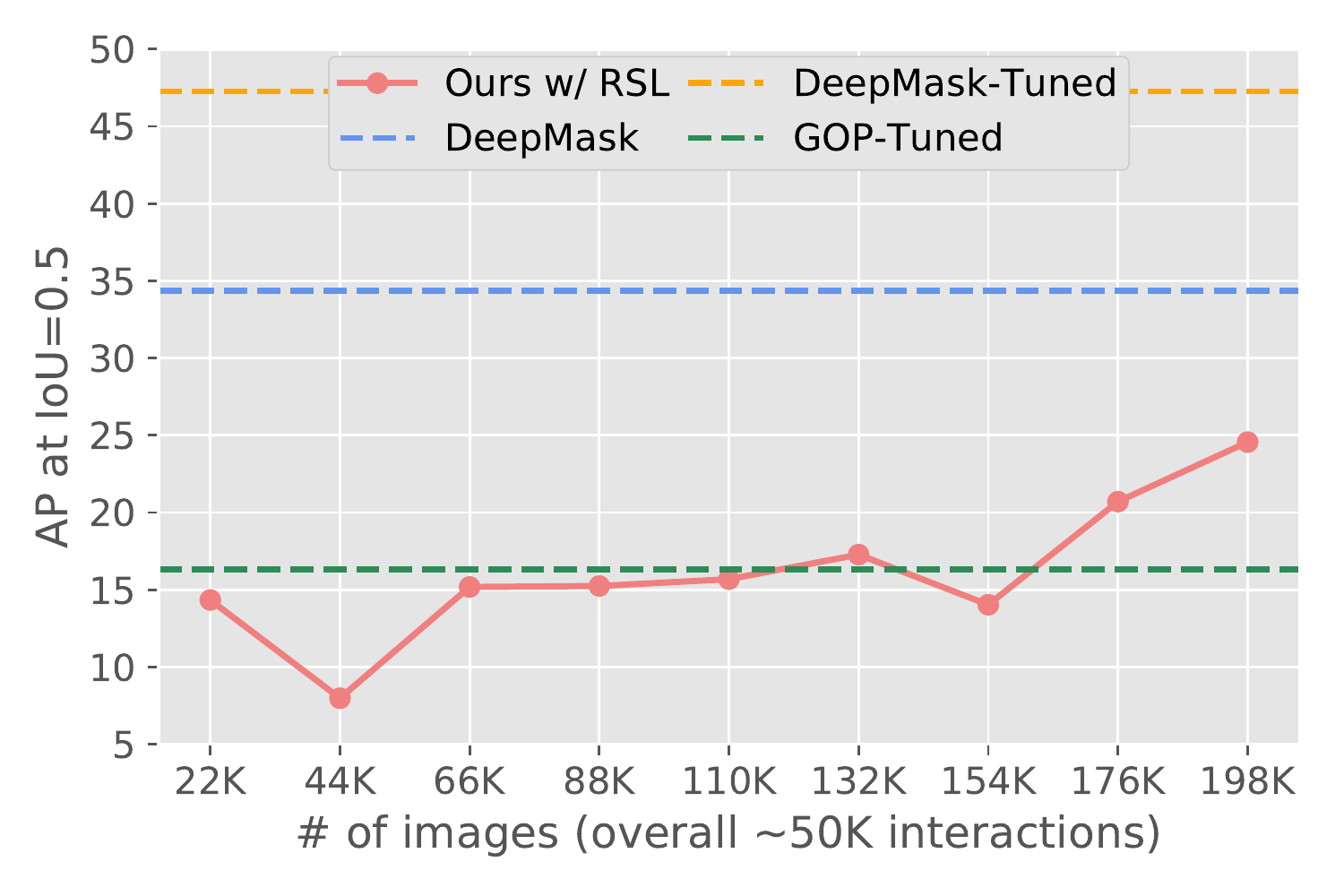}}
\subfigure[\smaller Successes vs. Interactions]{\label{fig:betterdata5}\includegraphics[width=0.32\textwidth]{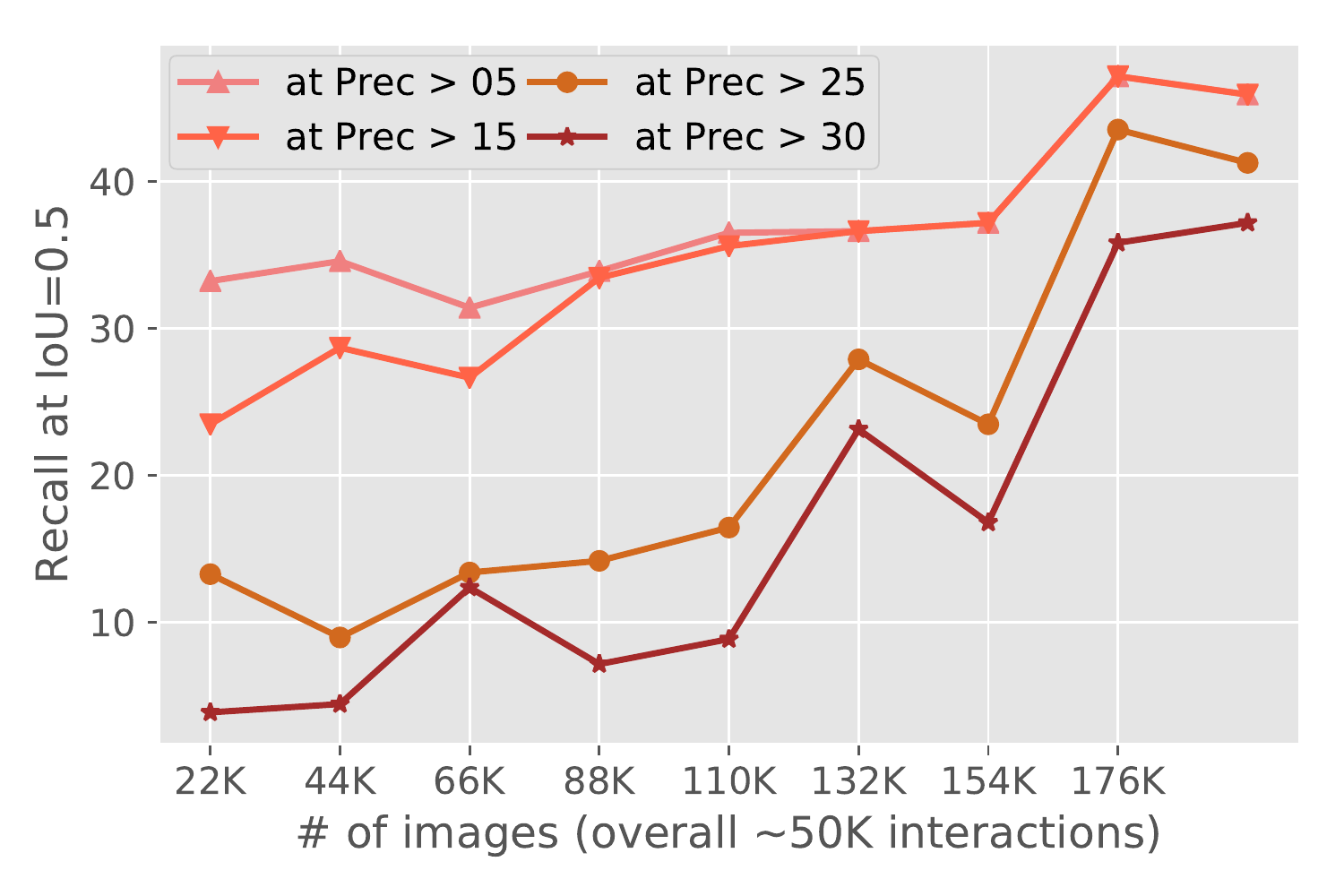}}
\subfigure[\smaller Precision vs. Recall]{\label{fig:precrec5}\includegraphics[width=0.32\textwidth]{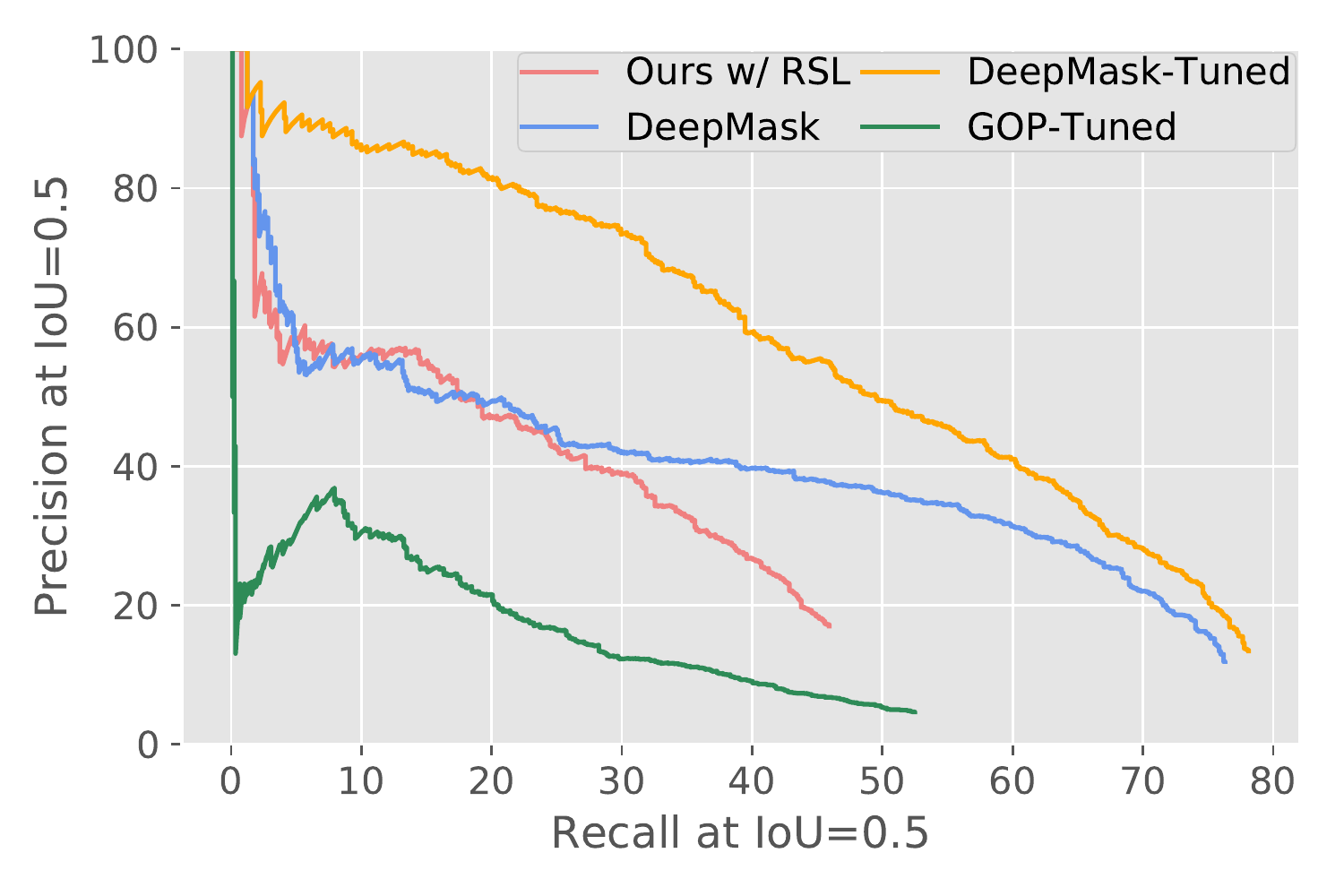}}
\caption{Quantitative evaluation of the segmentation model on the held-out test for AP at IoU of 0.5. (a) The performance of our system measured as mAP at IoU of 0.5 steadily increases with the amount of data. (b) The efficacy of experimentation performed by the robot is computed as the recall of ground truth objects that have IoU of more than 0.5 with the group of pixels that the robot believes to be objects. The steady increase in recall at different precision threshold shows that the robot learns to perform more efficient experiments with time. (c) Precision-Recall curves re-confirm the results.}
\label{fig:performance5}
\end{figure}

AP evaluation at IOU 0.5 (see Figure~\ref{fig:performance5}) reveals that while our method outperforms GOP, it is outperformed by DeepMask. We believe the main reason is that the masks obtained by robot interaction for training the segmentation model are imperfect (Section 4.1). Because of these imperfections, it is natural to expect that (a) performance will be worse at a more strict criterion (i.e. IOU 0.5) on mask quality; and (b) more number of noisy data points will be required to compensate for the noise and obtain higher quality masks. Just like the curve shown in Figure-4 of main paper for IoU 0.3, the performance of our system measured at IOU 0.5 is steadily increasing and is expected to catch up with DeepMask performance.

\begin{figure}[ht!]
\centering
\subfigure[Train Set]{\label{fig:trainobj}\includegraphics[width=0.32\linewidth]{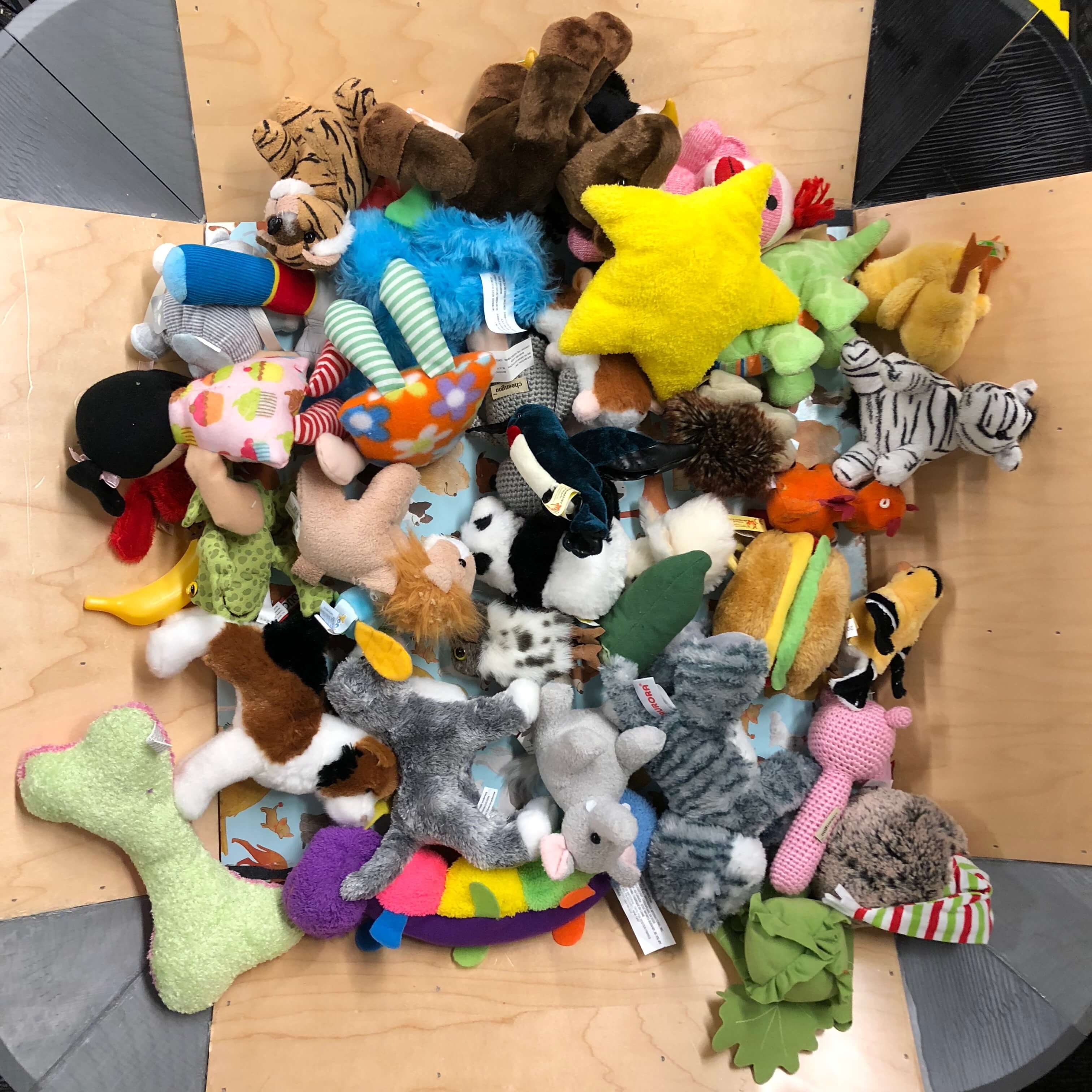}}
\subfigure[Val Set]{\label{fig:valobj}\includegraphics[width=0.32\linewidth]{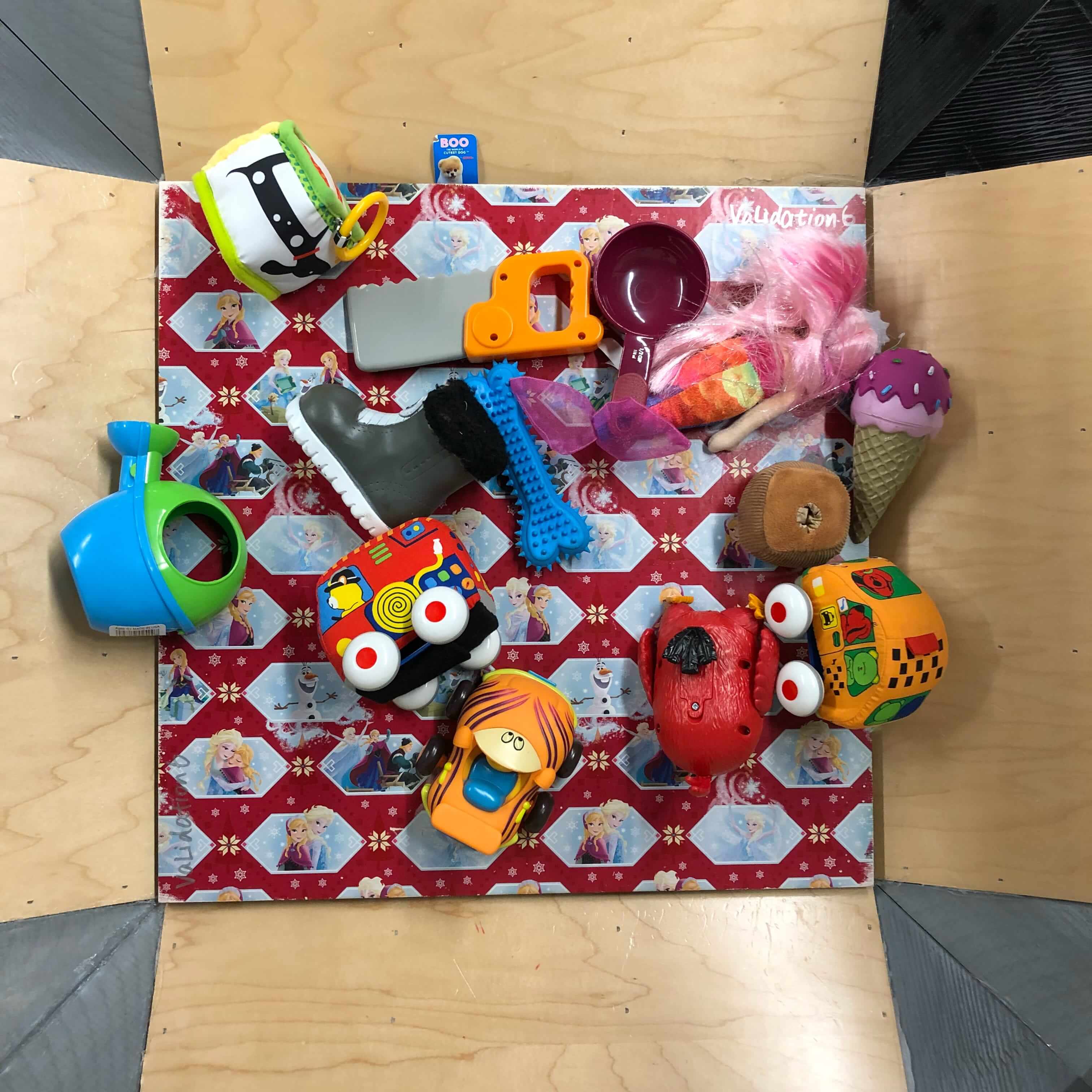}}
\subfigure[Test Set]{\label{fig:testobj}\includegraphics[width=0.32\linewidth]{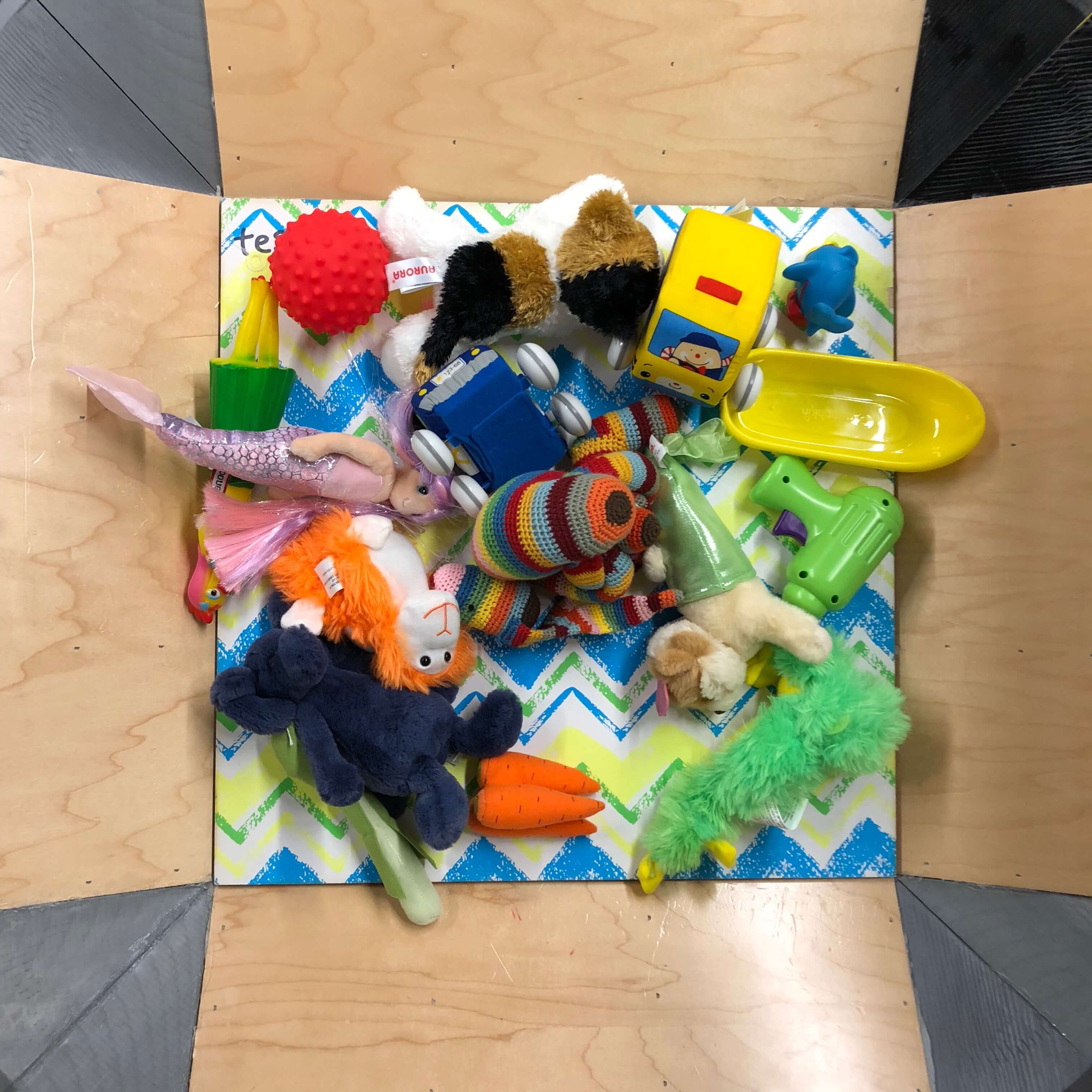}}
\caption{Visualization of the set of objects used for training (33), validation (13), and testing (16). Separate sets of backgrounds were used for training, validation and testing.}
\vspace{-4mm}
\label{fig:val}
\end{figure}

\begin{figure*}[ht!]
\centering
\subfigure[Training Backgrounds]{\label{fig:ap3A}\includegraphics[width=\textwidth]{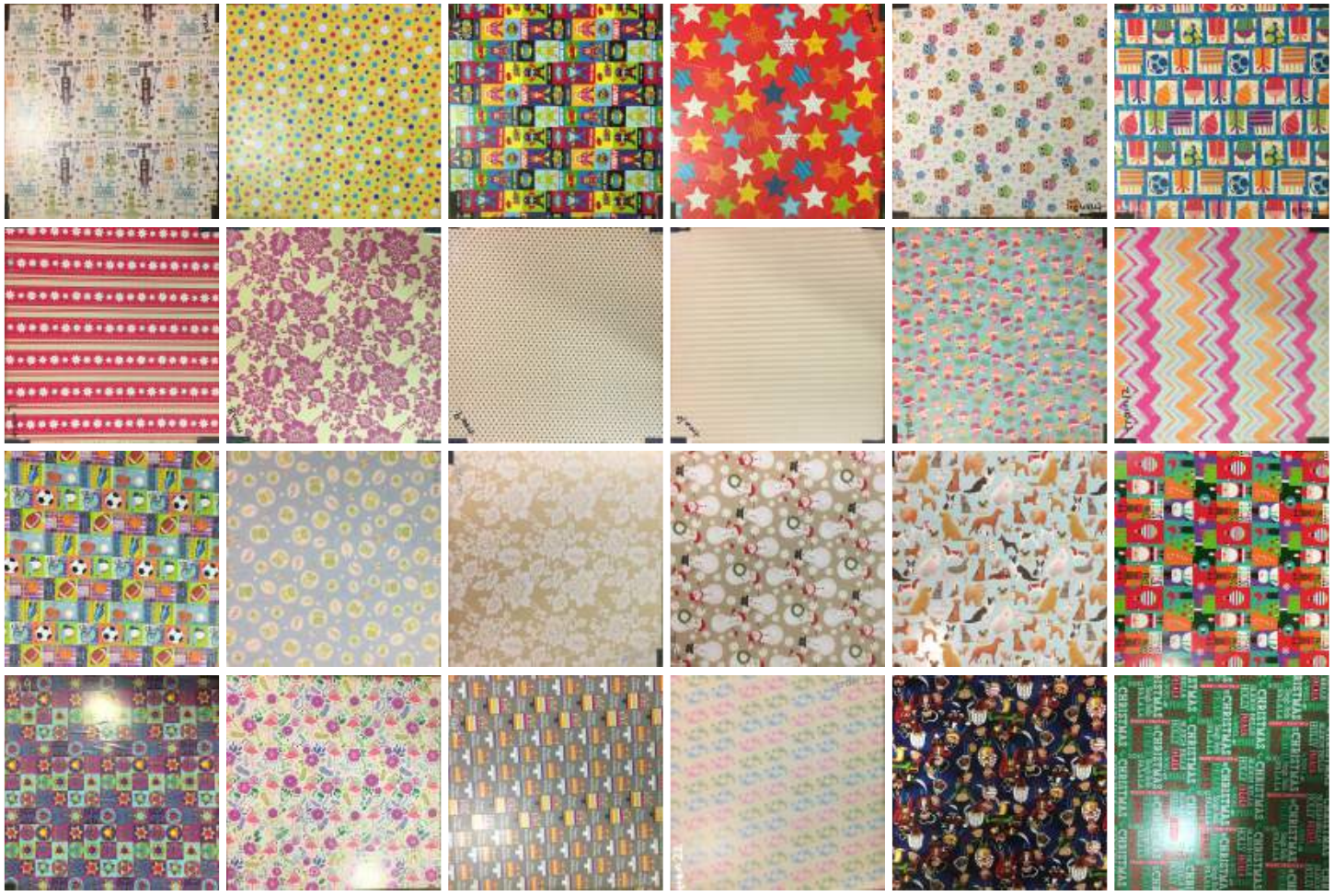}}
\subfigure[Validation Backgrounds]{\label{fig:ap3B}\includegraphics[width=\textwidth]{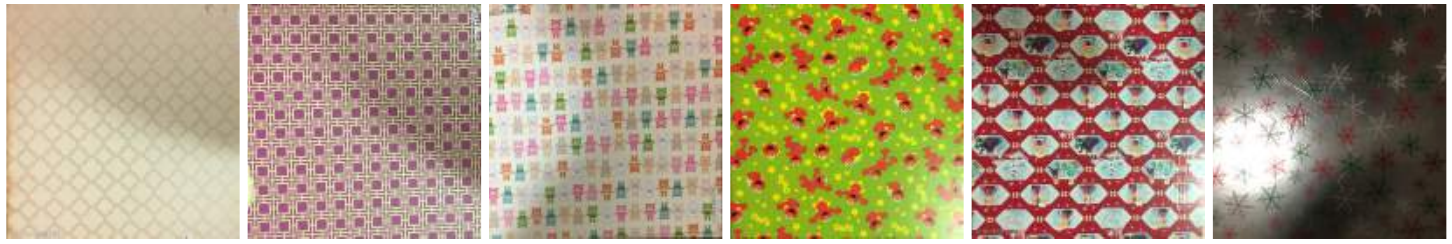}}
\subfigure[Testing Backgrounds]{\label{fig:ap3C}\includegraphics[width=\textwidth]{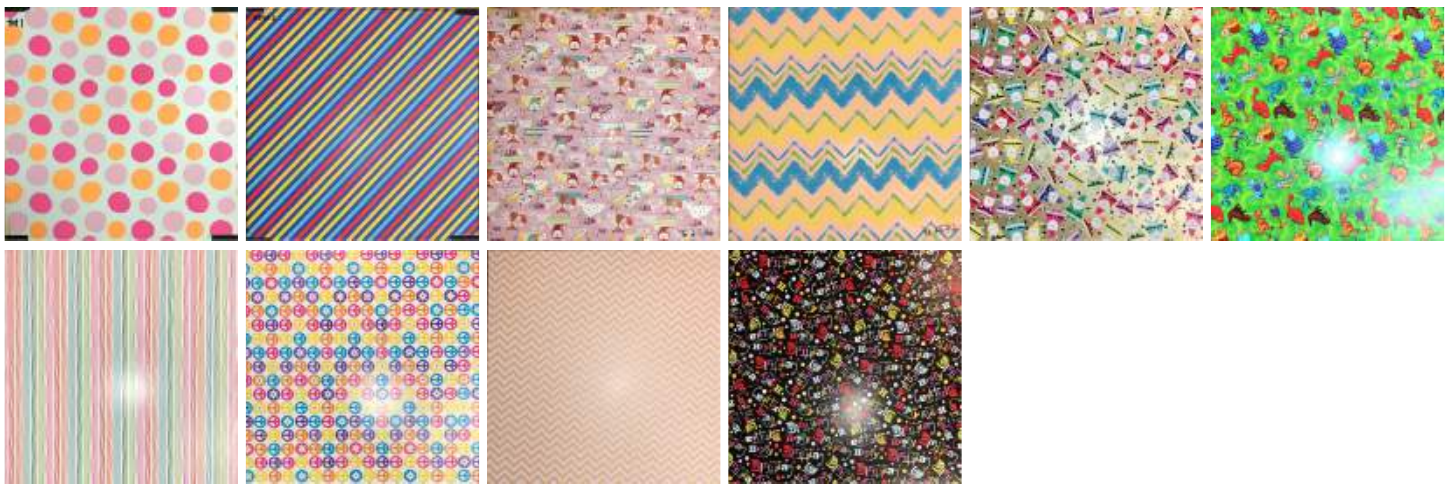}}
\caption{Visualization of backgrounds used for training, validation and test.}
\label{fig:backgrounds}
\end{figure*}

\end{document}